\documentclass[letterpaper, 10 pt, conference]{ieeeconf}  % Comment this line out if you need a4paper

\IEEEoverridecommandlockouts                              % This command is only needed if 
\usepackage{graphicx} % for pdf, bitmapped graphics files
\graphicspath{ {./images/} }
\usepackage{amsmath} % assumes amsmath package installed
\usepackage{amssymb}  % assumes amsmath package installed
\usepackage{booktabs}
\usepackage{pifont}
\usepackage{tabularx}
\usepackage{diagbox}
\usepackage{subfigure}
\usepackage{float}
\usepackage{tablefootnote}
\usepackage{hyperref}

\newcommand{\cmark}{\ding{51}}%
\newcommand{\xmark}{\ding{55}}%
\newcolumntype{Y}{>{\centering\arraybackslash}X}

\title{\LARGE \bf
NeRF-Loc: Visual Localization with Conditional Neural Radiance Field
}

\author{Jianlin Liu$^{1}$, Qiang Nie$^{1}$, Yong Liu$^{1}$ and Chengjie Wang$^{1}$ % <-this % stops a space
\thanks{$^{1}$Jianlin Liu, Qiang Nie, Yong Liu and Chengjie Wang are with Tencent, Shennan Boulevard, Nanshan District, Shenzhen, China {\tt\small \{jenningsliu, stephennie, choasliu, jasoncjwang\}@tencent.com}}%
}

\begin{document}

\maketitle
\thispagestyle{empty}
\pagestyle{empty}

%%%%%%%%%%%%%%%%%%%%%%%%%%%%%%%%%%%%%%%%%%%%%%%%%%%%%%%%%%%%%%%%%%%%%%%%%%%%%%%%
\begin{abstract}

We propose a novel visual re-localization method based on direct matching between the implicit 3D descriptors and the 2D image with transformer. A conditional neural radiance field(NeRF) is chosen as the 3D scene representation in our pipeline, which supports continuous 3D descriptors generation and neural rendering. By unifying the feature matching and the scene coordinate regression to the same framework, our model learns both generalizable knowledge and scene prior respectively during two training stages. Furthermore, to improve the localization robustness when domain gap exists between training and testing phases, we propose an appearance adaptation layer to explicitly align styles between the 3D model and the query image. Experiments show that our method achieves higher localization accuracy than other learning-based approaches on multiple benchmarks. Code is available at \url{https://github.com/JenningsL/nerf-loc}.

\end{abstract}

%%%%%%%%%%%%%%%%%%%%%%%%%%%%%%%%%%%%%%%%%%%%%%%%%%%%%%%%%%%%%%%%%%%%%%%%%%%%%%%%
\section{INTRODUCTION}

% 介绍camera-relocalization任务
Visual re-localization is the task of estimating camera's orientation and position in known scenes given a query image. It is an important module in simultaneous localization and mapping (SLAM) and structure from motion (SFM) systems, as well as being the prerequisite of various applications such as autonomous driving and augmented reality(AR).

% scene agnostic 的定位方法：matching
As the mainstream of solving visual localization problem, structure-based methods consist of two steps, 1) find correspondence between 3D points and 2D pixels 2) compute camera pose by PnP solver with Ransac. 3D-2D point pairs are typically obtained by scene agnostic feature matching.
% scene-specific 的定位方法：regression
In recent years, scene-specific localization methods attract increasing attention, where the scene is memorized in the neural network weights. Among these methods, direct pose regression\cite{kendall2015posenet}\cite{Sattler_2019_CVPR} has fast inference speed but low precision. Scene coordinate regression directly predicts absolute 3D coordinates of image pixels and uses scene structure explicitly to improve accuracy.
% 以前在定位里面用的3D模型一般是explicit的，比如点云。最近有的定位工作改成用implicit的3D模型，而implicit的3D模型本身也一般是scene-specific的
More recently, many efforts\cite{sucar2021imap}\cite{zhu2022nice}\cite{rosinol2022nerf}\cite{maggio2022loc} have been made to use implicit neural representation to replace explicit 3D models in localization pipeline. 
% 作为一种implicit 3D模型，NeRF有很多优点，但目前利用它做定位的研究还比较初步
Different from the commonly used discrete 3D models like point cloud and voxel grid, NeRF is one of the implicit 3D representations inferred from a sparse set of posed images, which models geometry and visual information in continuous 3D space. Photorealistic and differentiable rendering of NeRF enables many applications such as direct pose optimization and training set expansion. Nonetheless, existing methods using NeRF for pose estimation are either limited to data augmentation\cite{chen2021direct}\cite{chen2022dfnet}\cite{moreau2022lens} or analysis by synthesis\cite{yen2021inerf}. 
In these regards, we aim to directly localize the input image by matching it with a generalizable implicit neural 3D model. 

% \begin{table}
%   \label{table-0}
%   \caption{\textbf{Comparison with other methods.} Our method has the most desirable features for a visual localization pipeline. \footnotemark }
%   % footnote
%   \centering
%   \begin{tabularx}{\columnwidth}{c|Y|Y|Y|Y|Y}
   
%     \toprule
%     Method & Low-Texture & Multiple Scenes & Scene Prior & View Synthesis & Large Scene \\
%     \midrule
%     HLoc\cite{sarlin2019coarse}          & \xmark  & \cmark & \xmark  & \xmark   & \cmark \\
%     PixLoc\cite{sarlin2021back}        & \cmark  & \cmark & \xmark  & \xmark   & \cmark \\
%     Dsac\cite{brachmann2017dsac}          & \cmark  & \xmark & \cmark  & \xmark   & \xmark \\
%     Onepose\cite{sun2022onepose}       & \xmark  & \cmark & \xmark  & \xmark   & \xmark \\
%     iNeRF\cite{yen2021inerf}         & \cmark  & \xmark & \cmark  & \cmark   & \xmark \\
%     Ours          & \cmark  & \cmark & \cmark  & \cmark   & \cmark \\
%     \bottomrule
%   \end{tabularx}
% \end{table}

% \footnotetext{Low-Texture means being able to handle low-texture area. Multiple Scenes means that it can generalize to different scenes. Scene Prior refers to using scene-specific information. Novel View Synthesis ability enables direct model alignment by minimizing rendering loss.}

% 可泛化性和场景先验的矛盾
Unlike matching-based methods that generalize well across scenes, scene-specific localization methods are limited to the training scene, although they usually perform better under texture-less indoor conditions utilizing scene prior (per-scene information). Generalizability and scene prior seem to be contradictory in learning, how can we add scene prior to a localization model pretrained across multiple scenes? In this paper, we show that by re-designing the 3D representation in the visual localization pipeline, generalizability and scene prior can be both leveraged to attain better direct 3D-2D matching and localization accuracy. 

Inspired by the recent success of generalizable NeRF, our pipeline adopts a conditional NeRF model that can generate 3D features at arbitrary 3D locations. The generated 3D features are shared by both the subsequent matching and rendering. To keep the generalizability across scenes, our neural 3D model is conditioned on a support set that is composed of several posed reference images and depth maps. Based on the support set, continuous 3D descriptors are generated by aggregating multiview and point-wise features in a single forward pass. Similar to the training procedure of generalizable NeRF, our 3D model not only learns general matching during the joint training of multiple scenes as a good start but also memorizes coordinate-based scene prior in a residual way during per-scene optimization to boost the performance. With the learned conditional neural 3D model, efficient visual localization can be done by matching it with the image. To do this, some reference points are randomly sampled from any reconstructed 3D model to query 3D descriptors. Then, a transformer-based matcher is applied to estimate correspondences between the sampled 3D points and dense 2D pixels. When using our method with images in the wild, appearance changes between support images and target images are non-negligible for localization robustness and rendering quality. Therefore, we further propose an appearance adaptation layer to explicitly model the appearance factor in our 3D representation, which improves the localization robustness against domain changes. 

Our contributions are three-fold:
\begin{itemize}
    \item We propose a novel visual localization method based on matching between the conditional NeRF 3D model and 2D image, which formulates scene coordinate regression and feature matching in a unified framework. The proposed pipeline adopts a \textit{multi-scenes pretraining then per-scene finetuning} paradigm to learn both shared knowledge and scene prior for localization task.
    \item In order to tackle appearance changes between training support images and the query image, we propose an appearance adaptation layer to align image styles between the query image and the 3D model before matching. Experiments show that the robustness against domain changes is improved.
    \item Extensive experiments on real-world localization benchmarks are conducted to demonstrate the effectiveness of the proposed method. 
\end{itemize}

\section{Related Work}

\paragraph{Pose Estimation} 

% Image retrieval methods \cite{gordo2017end} \cite{arandjelovic2016netvlad} can provide rough pose estimation by finding the most similar reference images in the database. Sparsity of reference images limits the pose precision of image retrieval methods. To overcome this limitation, early work \cite{kendall2015posenet} directly regress global camera pose with CNN network. Several follow-up works improves \cite{kendall2015posenet} by adding photometric consistency loss \cite{chen2021direct} or augment training images by pose synthesis \cite{wu2017delving}. Besides absolute pose regression, relative pose regression methods\cite{balntas2018relocnet} \cite{ding2019camnet} predict the transformation between query image and the most relevant train image, which shows better accuracy and generalization. As pointed out in \cite{chen2020survey}, despite their scalability to large scene, most of the pose regression based methods fail to reach comparable performance over structure based methods. 

In the context of visual re-localization, 3D-2D matching is the essential problem. In recent years, deep learning has boosted the performance of 2D feature matching\cite{luo2019contextdesc}\cite{sarlin2020superglue}. To handle texture-less scene, detector-free 2D matching methods \cite{sun2021loftr}\cite{jiang2021cotr}\cite{chen2022aspanformer} has been proposed, which shows promising results. Acquiring 2D-3D correspondence from sparse 2D feature matching has its drawbacks, 1)\cite{germain2020s2dnet} pointed out that 2D keypoints may not reproduce under dramatic appearance changes, leading to failure of matching. 2)exhaustive matching between image pairs from retrieval is less efficient than direct 3D-2D matching \cite{sun2022onepose}. However, to the best of our knowledge, direct 3D-2D matching is seldom studied because cross-modality matching is difficult. Scene coordinate regression(SCR) takes a different path that directly regresses the dense 3D scene coordinates of input image. Dsac\cite{brachmann2017dsac} propose a differentiable counterpart of Ransac that enable end-to-end training of the scene coordinates based pose estimator. In the follow-up work, Dsac++\cite{brachmann2018learning} implements Dsac without learnable parameters which increases generalization capabilities. More recently, \cite{brachmann2021visual} extents Dsac++ to support RGB-D input and improve the initialization procedure. However, SCR is scene-specific because of per-scene training. To achieve scene agnostic localization, \cite{tang2021learning} proposes to construct a correlation tensor between query image and some reference images to regress coordinate map. Different from the existing methods, our method unifies scene coordinate regression to a direct 3D-2D matching framework, 3D-2D correspondences are established by matching 2D image features with 3D features from a conditional NeRF model. 

\paragraph{NeRF} NeRF is initially proposed in the seminal work \cite{mildenhall2020nerf} as an implicit 3D model for novel view synthesis. NeRF-W\cite{martin2021nerf} extends NeRF to images in the wild with varying appearances and transient objects. Since NeRF memorizes the scene in the network weights, it does not generalize to other scenes. To mitigate the burden of per-scene optimization, researchers have proposed several methods to build generalizable NeRF\cite{chen2021mvsnerf}\cite{liu2022neural}\cite{wang2021ibrnet}\cite{xu2022point}, where a general model is pretrained across different scenes and per-scene finetuning is applied later. In these methods, the NeRF model is conditioned on the local scene structure represented either as posed images or cost volume. We notice that scene-specific localization methods, such as scene coordinate regression, share the same idea with NeRF as they also remember the scene in neural network. In this work, we propose a modified version of conditional NeRF model to better suit visual localization pipeline.

\paragraph{NeRF+Pose Estimation.} As a 3D representation and differentiable renderer, NeRF can help visual localization by synthesizing more training images or refining pose with photometric loss. \cite{moreau2022lens} proposes an offline training data generation method based on NeRF to enhance camera pose regression performance. In \cite{yen2021inerf}, the authors propose to estimate pose by minimizing the difference between rendered image and query image. However, it is time-consuming and only validated on data without significant illumination changes. In \cite{chen2022dfnet}, a histogram-assisted NeRF is used to mitigate the domain gap between rendered image and real image, so that a better direct alignment loss can be used to train pose regression. Our method focuses on designing a more practical localization pipeline based on the conditional NeRF, where 3D-2D matching is utilized to avoid slow analysis by synthesis process while keeping all desirable properties brought by NeRF.

\section{Method}

\begin{figure*}[htbp]
  \centering
  \includegraphics[width=0.95\textwidth]{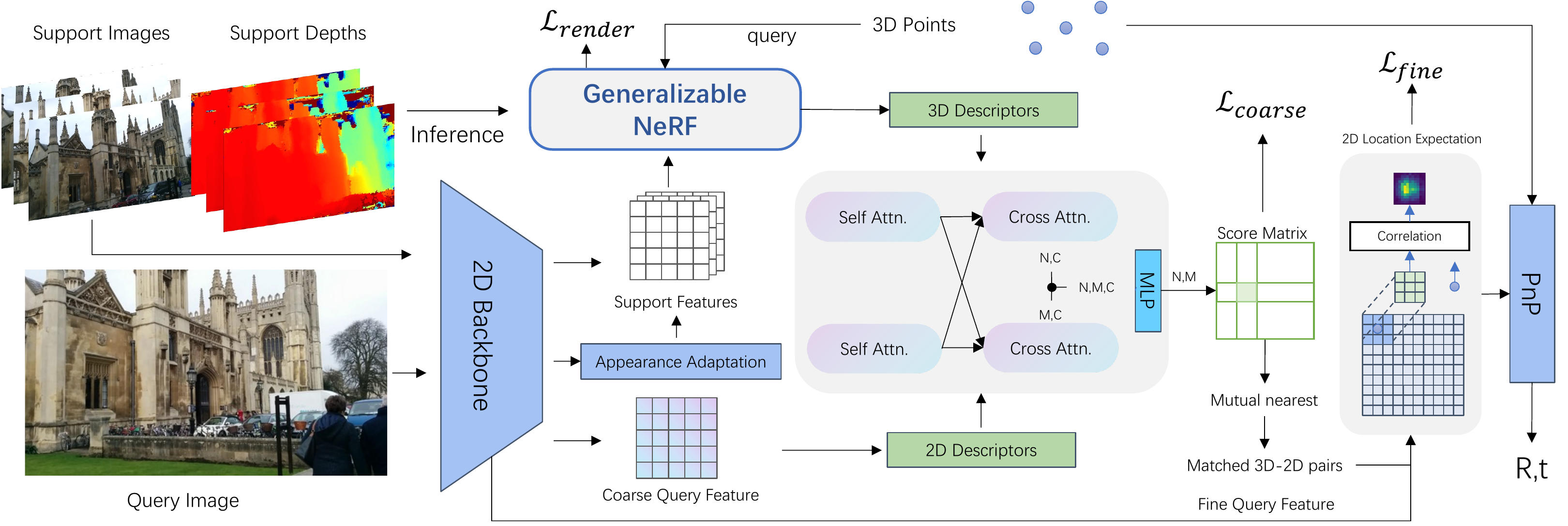}
  \caption{System overview. In our localization pipeline, the scene is represented as generalizable NeRF which is conditioned on support images. 3D points are fed into the NeRF model to generate 3D descriptors. 3D-2D correspondences are obtained by direct matching between 3D and 2D descriptors, in order to compute camera pose via PnP solver.}
  \label{fig-overview}
\end{figure*}

% 先介绍引入conditional NeRF作为3d模型的动机
We choose conditional NeRF as 3D model in our pipeline, the motivation behind is two-fold: 1) to facilitate more extension, such as direct 3D model alignment and expanding train set by synthesizing novel images. 2) to make use of the feature aggregation for better 3D descriptors.

To learn both general knowledge and scene prior, our method consists of two training phases. Firstly, the scene-agnostic pose estimator is trained across multiple scenes. Secondly, per-scene optimization is applied to further improve localization accuracy. Specifically, the pose estimator is realized by predicting the 2D projection of 3D model points. Inspired by LoFTR\cite{sun2021loftr}, a coarse-to-fine matching scheme is adopted to directly estimate 3D-2D correspondences. Note that LoFTR performs dense-to-dense matching between image pair, while our method performs sparse-to-dense matching between 3D model and image. Since matching all 3D model points is computationally expensive and even infeasible sometimes, 3D keypoint sampling is applied to keep only a small portion of points before matching. It's worth mentioning that this sparse-to-dense matching avoids the potential poor repeatability of 2D keypoint detector under large viewpoint changes. After the 3D-2D matches are established, a basic PnP solver with Ransac is used to compute camera pose.

\subsection{Conditional Neural 3D Model}
\label{sec-3d_model}
% 基于multiview RGBD，image blending(neuray 遮挡推理) + pointnerf
In this section, we will introduce the 3D model representation in our method. To perform direct 3D-2D matching, the 3D model predicts the associated 3D descriptor for a given 3D location $x$ as query. To facilitate more downstream tasks, a generalizable NeRF is chosen as the 3D model which is shared by the matching module and neural rendering module. As generalizability is desirable for localization, our method constructs a NeRF model on the fly, conditioned on several support images with known pose. While building a NeRF model in one forward pass is difficult. To make the problem easier, a noisy and incomplete depth map for each support frame is assumed to be available beforehand. Unlike \cite{xu2022point} where source depth maps are generated at inference time, we generate these support depth maps offline with off-the-shelf MVS method. There are two benefits of this: 1) MVS time can be saved during inference, 2) It is compatible with depth maps from range sensors. 
Before computing 3D descriptors for query points, the support RGBD images need some processing to better represent the local scene. Given the support images and depths, image features are first extracted by a 2D backbone. Then, with known camera parameters, the image feature and raw depth are lifted to 3D neural support points $P = \{P_{s}, F_{s}, \Lambda_{s}, D_{s}\}$ as in \cite{xu2022point}, where $P_{s}$ $F_{s}$ $\Lambda_{s}$ $D_{s}$ represent location, feature, confidence and viewing direction of support points respectively. Inspired by \cite{liu2022neural}, reference depth maps can also be used for visibility reasoning though they are noisy and incomplete. The raw depths are warped and fused in each support frame to be utilized by a visibility reasoner. The visibility reasoner estimates the likelihood of whether a certain 3D point is visible in a specific reference frame, which requires a complete depth of that reference frame. Hence, this visibility reasoning can be considered as a cross-frame depth validation and completion step. For more details, please refer to \cite{liu2022neural}.

In order to compute the 3D feature of query points $X\in \mathbb{R}^{N \times 3}$, we combine the projected multiview features and the neural support points in a complementary way. Specifically, $X$ are projected to each support frame for fetching multiview features, which are aggregated by computing the visibility weighted mean and variance. The visibility-aware aggregated multiview features $F_{m} \in \mathbb{R}^{N \times C}$ contain the geometry and visual information at location $X$, which are equivalent to the elements of MVS cost volume. To augment $F_{m}$ with local scene structure, we use KNN to search k-nearest neighbors of $X$ in neural support points $P$ and take their point features $F_{s}\in \mathbb{R}^{N \times K \times C}$. $F_{m}$ and $F_{s}$ are passed into a multi-heads attention(MHA) block to get correlated support point feature $F^{'}\in \mathbb{R}^{N \times K \times C}$. In this way, multiview features and support point features interact with each other to get better feature aggregation. The MHA module also outputs the attention weights $W_a$ for each local support point, which are multiplied by the inverse distance weights $W_d$ and confidence $\Lambda_{s}$ of local support points to get the final local weights $W=W_a*W_d*\Lambda_{s}\in \mathbb{R}^{N \times K}$. Finally, the 3D feature for $X$ is calculated as,
\begin{equation}
\label{eq:3d_feature}
\mathcal{M}(X)= \{f(x_i)=\sum_{k=0}^{K-1}\frac{w_{ik}*f_{ik}^{'}}{K}; x_i \in X , w_i \in W, f_i^{'} \in F^{'}\}
\end{equation}

\begin{figure*}[h]
  \centering
  \includegraphics[width=\textwidth]{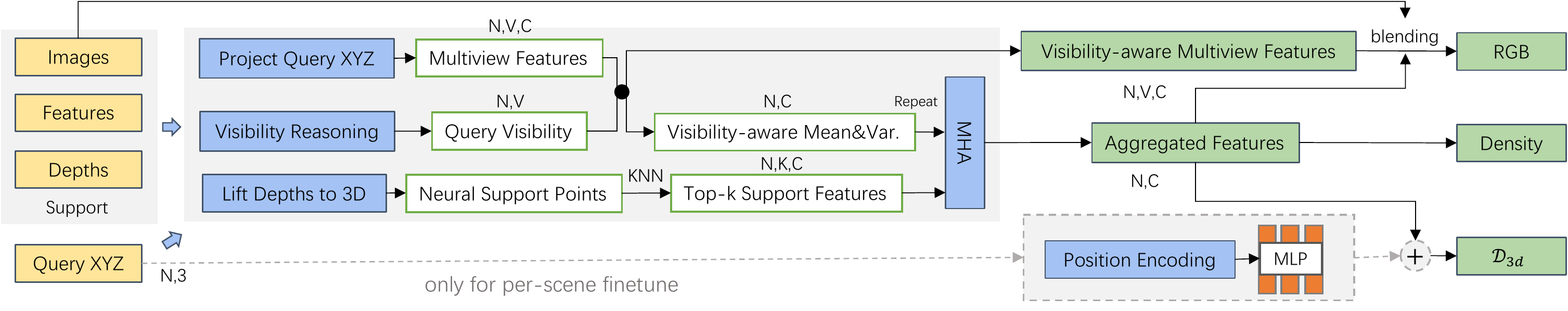}
  \caption{Architecture of our conditional NeRF model, a feature generator for any 3D location is shared by novel view synthesis and 3D-2D matching. V refers to the number of support images. For more details about visibility reasoning, please refers to \cite{liu2022neural}.}
  \label{fig-arch}
\end{figure*}

% \begin{figure}[h]
% %   \centering
%   \includegraphics[width=\columnwidth]{aggregation}
%   \caption{Aggregation in 3D model. Our 3D model is conditioned on both support images and the back-projected 3D neural points.}
%   \label{fig-aggregation}
% \end{figure}

% 这里介绍 coordinate-based feature 是怎么加入的
$\mathcal{M}(X)$ is scene agnostic 3D representation since it is computed from the support images, It works well for multi-scenes localization training. However, it does not utilize any scene prior explicitly for finding 3D-2D correspondences, which is valuable when per-scene finetuning is available. To this end, we propose a simple yet effective augmentation for this scene agnostic 3D representation. As indicated in Fig.\ref{fig-arch}, the 3D coordinate of query points are passed to position encoding followed by an MLP to produce coordinate-based features. These coordinate-based features are added to the scene-agnostic 3D descriptors as residuals. Note that this operation is only used for per-scene optimization stage. We empirically found that scene prior features further boost the localization performance significantly.

\subsection{Appearance Adaptation Layer}

\begin{figure}[h]
%   \centering
  \includegraphics[width=\columnwidth]{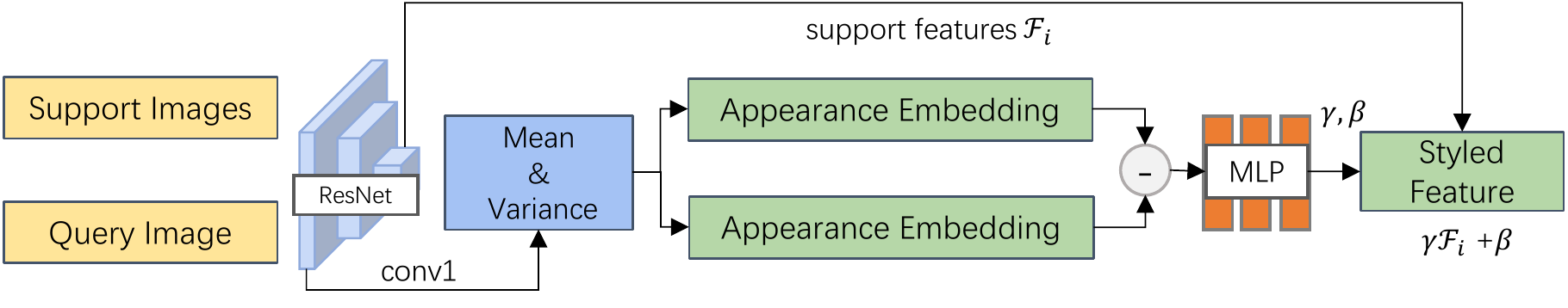}
  \caption{Appearance adaptation.}
  \label{fig-appearance}
\end{figure}
Our 3D model depends on support images, which may have different lighting conditions or exposures to the query image. The 3D-2D matching will deteriorate when the style of the query image changes. Other methods \cite{sarlin2021back} usually tackle this problem by using more diverse training images to learn invariant features. Though the high-level features are usually quite robust to appearance changes, eliminating domain gaps can further improve the robustness. On the other hand, novel view synthesizer will have difficulty in rendering image that is similar to the target image, which leads to bad performance on direct model alignment. NeRF-W \cite{martin2021nerf} proposes to learn the appearance embedding as parameters for each training image individually. However, for visual localization problem, the target appearance embedding should be computed from the query image. To obtain better robustness against low-level image statistics changes, we propose an appearance adaption layer that explicitly aligns the style of support images and target image. As shown in Fig.~\ref{fig-appearance}, we extract appearance embedding from both the target image and support images. Image feature pyramid is defined as $\mathcal{F}_i$, where $\mathcal{F}_0$ refers to the original RGB image. Following \cite{huang2017arbitrary}, channel-wise mean and variance of a fixed low-level feature map $\mathcal{F}_1$ is used as appearance embedding. Appearance embedding of the target image is subtracted from those embeddings of support images and then fed into an MLP to get appearance difference embedding $\delta$. Given a specific feature level $\mathcal{F}_i \in \mathbb{R}^{H\times W\times C}$ of support images to be modulated, $\delta$ is decoded into a scaling factor $\gamma$ and offset $\beta$ of $C$ channels. Finally, the global affine transformation defined by $\gamma$ and $\beta$ is applied to $\mathcal{F}_i$ so that it is aligned with the target style, which is formulated as $AD(\mathcal{F}_i)=\gamma \mathcal{F}_i + \beta$.
Appearance adaptation layer is applied to modulate $\mathcal{F}_0$, $\mathcal{F}_2$ and $\mathcal{F}_3$ respectively.

\subsection{Localization Pipeline}
In this section, we will introduce our localization pipeline based on the conditional neural 3D model introduced in section \ref{sec-3d_model}. Overall, our localization system consists of three steps: support images selection, 3D-2D matching and pose estimation by solving PnP.

% support images的构建方式
As the first step, how to construct support set depends on the application scenario. We utilize two methods that will cover most of the applications. 
\subsubsection{Image retrieval} A general way is using image retrieval to select support frames. This method works well for the outdoor scene, which makes our method applicable to very large scene. Despite its simplicity and effectiveness, image retrieval performs poorly in indoor scenes with repetitive patterns and object-level scenes, and does not guarantee the spatial uniformity of the selected images. 
\subsubsection{Image coreset} For small scenes such as an object instance, an evenly down-sampled trajectory can be used as support images, which is named image coreset. To attain image coreset, we apply \textit{farthest pose sampling} on all training frames, which repeatedly picks the next frame that has the most different viewing angle from any selected frame, until the max number of frames is reached.
% coarse to fine 的 3D-2D matching. 构建两层的neural 3D model；cascade的matching方式；

After the support images are selected, a coarse to fine transformer-based matching module is leveraged to find 3D-2D correspondences directly. Specifically, we extract coarse and fine 3D descriptors from the neural 3D model to represent the scene geometry.
%we use two level neural 3D models to represent the scene geometry, from whom coarse and fine 3D descriptors are extracted. 
These two level 3D models are denoted as $\mathcal{M}_c$ and $\mathcal{M}_f$, where $\mathcal{M}_c$ is based on $\mathcal{F}_0$ and $\mathcal{F}_3$, $\mathcal{M}_f$ uses $\mathcal{F}_0$ and $\mathcal{F}_2$. 

In coarse level matching, 3D reference points $X\in \mathbb{R}^{N\times 3}$ are randomly sampled and fed into $ \mathcal{M}_c $ to get 3D descriptors $\mathcal{D}_{3d} = \mathcal{M}_{c}(X) \in \mathbb{R}^{N\times C}$. Note that $X$ can be sampled from any 3D point cloud such as the sparse points from SFM, as long as it reveals the correct scene structure. However, we choose to simply sample from neural support points. The coarse level target feature map is reshaped to get 2D descriptors $\mathcal{D}_{2d} \in \mathbb{R}^{HW\times C}$ for coarse matching. After that, $\mathcal{D}_{3d}$ and $\mathcal{D}_{2d}$ are transformed to $\mathcal{D}^{'}_{3d}$ and $\mathcal{D}^{'}_{2d}$ by cross and self-attention layers. The transformed descriptors are used to compute correlation tensor of shape $(N,HW,C)$ which is converted to correlation matrix of shape $(N,HW)$ with MLP later. Therefore, the coarse level matching score $\mathcal{S}$ is formulated as $\mathcal{S} = sigmoid(mlp(\mathcal{D}^{'}_{3d}\odot\mathcal{D}^{'}_{2d})) \in \mathbb{R}^{N\times HW}$, where $\odot$ refers to computing correlation tensor, as shown in Fig.\ref{fig-overview}. Based on $\mathcal{S}$, coarse 3D-2D correspondences are generated by filtering with predefined score threshold $\tau$ and mutual nearest checking. To train the coarse matcher, we compute the ground truth score matrix $\mathcal{S}^{*}$ with the provided depth map. Focal loss\cite{lin2017focal} is adopted as the coarse matching loss function.

% \begin{equation}
% \label{eq:coarse_loss}
%     \begin{aligned}
%     \mathcal{L}_{coarse}&= \sum\limits_{p_{(i,j)}\in \mathcal{S}}\frac{-(1-p_{(i,j)})^{\lambda}log(p_{(i,j)})}{N\times HW}, \\
%     where, p_{(i,j)}&=
%     \begin{cases}
%         \mathcal{S}_{(i,j)},   & \text{if $\mathcal{S^{*}}_{(i,j)}$=1} \\
%         1-\mathcal{S}_{(i,j)}, & \text{otherwise}
%     \end{cases}
%     \end{aligned}
% \end{equation}

In fine level matching, for each matched 3D-2D pair from coarse matching, we construct the fine level 2D descriptors $\mathcal{D}_{2d}^{fine}\in \mathbb{R}^{M\times 7\times 7 \times C}$ by taking a feature patch centered at the coarse 2D location, and query fine level 3D descriptors $\mathcal{D}^{fine}_{3d} = \mathcal{M}_{f}(X_{matched}) \in \mathbb{R}^{M\times C}$, where $M$ is the number of valid coarse matches. Similar with the coarse matching, $\mathcal{D}_{3d}^{fine}$ and $\mathcal{D}_{2d}^{fine}$ are first transformed by a self/cross-attention layers. Then a correlation matrix is computed by $\mathcal{S}^{fine}=softmax(mlp(\mathcal{D}_{3d}^{fine'}\odot\mathcal{D}_{2d}^{fine'}))$. Finally, 2D coordinates are refined to sub-pixel level by adding spatial expectation $\hat{\mathcal{E}}$ based on $\mathcal{S}^{fine}$. Following LoFTR, the fine level matching is supervised by L2 loss between predicted and ground truth location $\mathcal{L}_{fine}=||\hat{\mathcal{E}}-\mathcal{E}||^2$. For more detail, please refer to \cite{sun2021loftr}. 

To keep the benefits of NeRF as 3D model, an auxiliary rendering head is added upon the shared 3D feature from $\mathcal{M}_f$, which is supervised by L2 loss $\mathcal{L}_{render}$. Following \cite{liu2022neural}, depth recovering loss $\mathcal{L}_{depth}$ is added to provide direct supervision on visibility reasoning, where the network is required to remove noises from support depth maps during training. In summary, the final training loss is defined as followed, 
\begin{equation}
\mathcal{L} = \mathcal{L}_{coarse}+\mathcal{L}_{fine}+\mathcal{L}_{render}+\mathcal{L}_{depth}
\end{equation}

\section{Experiments}

\subsection{Datasets}
We evaluate our pipeline on four localization benchmarks, ranging from object level to outdoor scene.
\paragraph{Indoor} 12Scenes\cite{valentin2016learning} and 7Scenes\cite{shotton2013scene} are two room-size indoor localization datasets, where depth images are scanned by the structure light sensor and the ground truth poses are provided. There are no significant illumination changes or dynamic objects. 12Scenes contains 12 scenes and each contains several hundreds of frames. Each scene in 7Scenes has several thousands of frames. RGB images are registered to depth maps in 12Scenes but not in 7Scenes. Following \cite{brachmann2021visual}, we register the images in 7Scenes first. Camera parameters of 7Scenes are less accurate, which brings challenges to fitting a global consistent 3D model.

\paragraph{Outdoor} Cambridge Landmarks\cite{kendall2015posenet} dataset scaling from $875m^2$ to $5600m^2$ is for evaluating localization algorithms in large-scale outdoor scenes. Illumination and exposure of camera are different between sequences. Transient objects commonly exist in this dataset.

\paragraph{Object} Onepose\cite{sun2022onepose} is an object-level pose estimation dataset collected by hand-held smartphones. Several videos are captured around object instances with different backgrounds. Trajectories of different videos are aligned to a unified coordinate system.

\subsection{Evaluation Protocol}
Localization accuracy is defined as the ratio of correctly localized frames, given a criterion of success. For all datasets except Cambridge, a localized frame with rotation error below 5° and translation error below 5cm is considered as correct. As the scene scale of Cambridge dataset varies a lot, the translation threshold should be adaptively set for each scene. Here, the same translation thresholds with \cite{brachmann2021visual} are used (35cm for St. Mary’s Church, 45cm for Great Court, 22cm for Old Hospital, 38cm for King’s College and 15cm for Shop Facade). Besides localization accuracy, the commonly used median translation(cm) and rotation(°) error are also reported on Cambridge dataset. The same train/test split with \cite{brachmann2021visual} is used.

\subsection{Implementation Details}

For each dataset, the network is pretrained across all scenes for 30 epochs, and then optimized per scene for another 30 epochs, using Adam optimizer with learning rate of $5*10^{-4}$. Geometry augmentations (\textit{random zooming} and \textit{random rotation}) and Color jitter are applied to input images. Appearance adaptation is not used for Onepose, since the images are cropped by 2D detection box and padded with black background. To select support images, Image retrieval\cite{arandjelovic2016netvlad} is used for all datasets except Onepose, which uses image coreset. For image retrieval, 5 images are randomly selected from the top 20 retrieval images during training, while top-10 images are used during test time. For image coreset, 16 support images are kept for both training and testing. We use the MVS algorithm in Colmap to prepare depth maps for training images except for the indoor datasets. For matching in all experiments, 1024 3D points are randomly sampled. Localizing one frame takes 250ms on Nvidia V100 GPU, using 10 support images.

\subsection{Comparison with Other Methods}

\begin{figure*}
\setkeys{Gin}{width=\linewidth}
\setlength{\tabcolsep}{0pt}
\renewcommand{\arraystretch}{1}
\begin{tabularx}{\textwidth}{XXXXXXX}
  \includegraphics{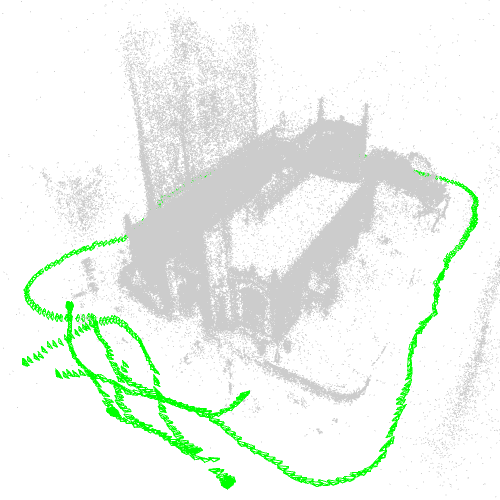} &   \includegraphics{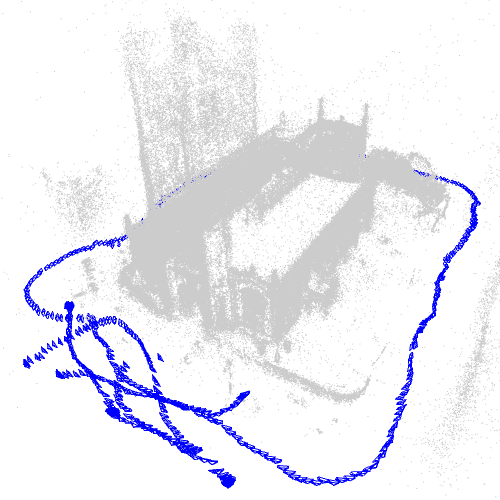} & \includegraphics{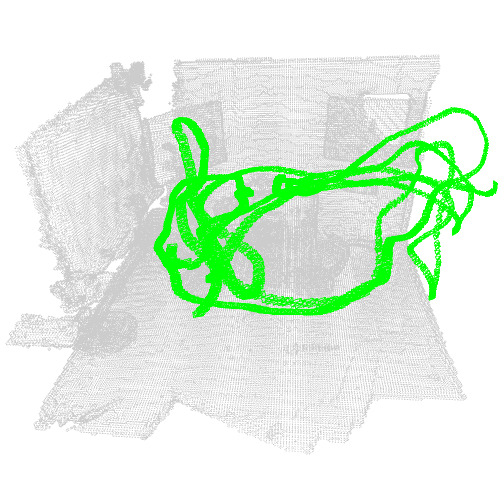} & \includegraphics{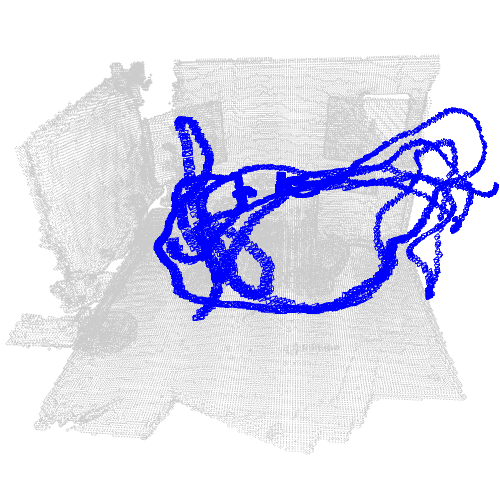} & \includegraphics{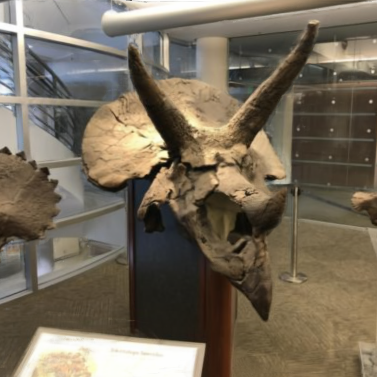} & \includegraphics{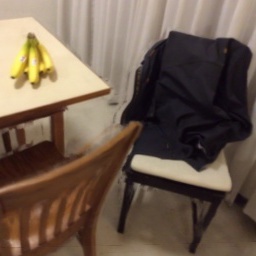} & \includegraphics{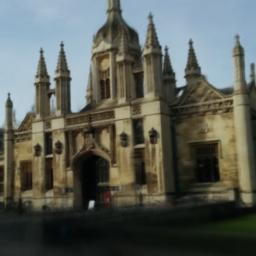} \\
\end{tabularx}
\caption{Qualitative results of pose estimation(first four, Green: groundtruth, Blue: prediction) and rendering(last three). }
\label{fig-qualitative-result}
\end{figure*}

As indicated in Table~\ref{table-1}, \ref{table-indoor} and \ref{table-onepose}, our method consistently achieves the best overall performance among learning-based visual localization methods on all four datasets, which shows the practical value of our system. Qualitative results can be found in Figure \ref{fig-qualitative-result} and the supplementary video. %Qualitative results and per-scene accuracy can be found in Appendix.

\begin{table*}[htb]
  \caption{\textbf{Evaluation on outdoor and indoor localization benchmarks.} We report the median translation(cm) and rotation(°) error for each scene. \textbf{Avg.} stands for the average median error, and \textbf{Acc.} is the abbreviation for average accuracy. }
  \label{table-1}
  \centering
  \setlength{\tabcolsep}{4pt}
  \begin{tabular}{lllllllllllllll}
    \toprule
    &\multicolumn{6}{c}{Cambridge Landmarks - outdoor}  &\multicolumn{8}{c}{7scenes - indoor}                 \\
    \cmidrule(r){2-7}
    \cmidrule(r){8-15}
    Method & Church & Court & Hospital & College & Shop & Avg.$\downarrow$ & Chess & Fire & Heads & Office & Pumpkin & Kitchen & Stairs & Acc.\\
    \midrule
    SANet\cite{yang2019sanet} & 16/0.57 & 328/1.95 & 32/0.53 & 32/0.54 & 10/0.47 & 83.6/0.8   & 3/0.9 & 3/1.1 & 2/1.5 & 3/1.0 & 5/1.3 & 4/1.4 & 16/4.6 & 68.2 \\
    DSAC\cite{brachmann2017dsac} & 55/1.6 & 280/1.5 & 33/0.6 & 30/0.5 & 9/0.4 & 81.4/0.9 & 2/0.7 & 3/1.0 & 2/1.3 & 3/1.0 & 5/1.3 & 5/1.5 & 190/49.4 & 60.2 \\
    InLoc\cite{taira2018inloc} & 18/0.6 & 120/0.6 & 48/1.0 & 46/0.8 & 11/0.5 & 48.6/0.7   & 3/1.1 & 3/1.1 & 2/1.2 & 3/1.1 & 5/1.6 & 4/1.3 & 9/2.5 & 66.3 \\
    DSM\cite{tang2021learning} & 12/0.4 & 44/0.2 & 24/0.4 & 19/0.4 & 7/0.4 & 21.2/0.4 & 2/0.7 & 2/0.9 & 1/0.8 & 3/0.8 & 4/1.2 & 4/1.2 & 5/1.4 & 78.1 \\
    DSAC++\cite{brachmann2018learning} & 13/0.4 & 40/0.2 & 20/0.3 & 18/0.3 & 6/0.3 & 19.4/0.3 & \textbf{2/0.5} & 2/0.9 & 1/0.8 & 3/0.7 & 4/1.1 & 4/1.1 & 9/2.6 & 74.4 \\
    DSAC*\cite{brachmann2021visual} & 13/0.4 & 49/0.3 & 21/0.4 & 15/0.3 & 5/0.3 & 20.6/0.3   & 2/1.1 & 2/1.2 & 1/1.8 & 3/1.2 & 4/1.3 & 4/1.7 & 3/1.2 & 85.2\\
    % NG-DSAC++ & 10/0.3 & 35/0.2 & 22/0.4 & 13/0.2 & 6/0.3 & 17.2/0.3 \\
    HACNet\cite{li2020hierarchical} & 9/0.3 & 28/0.2 & 19/0.3 & 18/0.3 & 6/0.3 & 16.0/0.3  & 2/0.7 & 2/0.9 & 1/0.9 & 3/0.8 & 4/1.0 & 4/1.2 & 3/0.8 & 84.8\\
    PixLoc\cite{sarlin2021back} & 10/0.3 & 30/0.1 & \textbf{16/0.3} & 14/0.2 & 5/0.2 & 15/0.2   & 2/0.8 & \textbf{2/0.7} & \textbf{1/0.8} & 3/0.8 & 4/1.2 & \textbf{3/1.2} & 5/1.3 & 75.7 \\ 
    \midrule
    Ours & \textbf{7/0.2} & \textbf{25/0.1} & 18/0.4 & \textbf{11/0.2} & \textbf{4/0.2} & \textbf{13/0.2} & 2/1.1 & 2/1.1 & 1/1.9 & \textbf{2/1.1} & \textbf{3/1.3} & 3/1.5 & \textbf{3/1.3} & \textbf{89.5}\\
    \bottomrule
  \end{tabular}
\end{table*}

\begin{table}[H]
  \caption{\textbf{Localization accuracy on two indoor datasets.} Results of other methods are from \cite{brachmann2021visual}.}
  \label{table-indoor}
  \centering
  \begin{tabularx}{\columnwidth}{c||cXcXXX}
    \toprule
    Method     & ORB+PnP & DSAC & DSAC++ & DSAC* & SCoCR & \textbf{Ours} \\
    \midrule
    12Scenes & 53.7 & 83.5 & 96.8 & 99.1 & 99.3 & \textbf{99.8}\\
    7scenes  & 40.7 & 60.2 & 74.4 & 85.2 & 84.8 & \textbf{89.5}\\
    \bottomrule
  \end{tabularx}
\end{table}

\begin{table}[H]
  \caption{\textbf{Localization accuracy on object-level dataset Onepose.}}
  \label{table-onepose}
  \centering
  \begin{tabularx}{\columnwidth}{c||YYYYYYY}
    \toprule
    Method & 0447 & 450 & 0488 & 0493 & 0494 & 0594 & Avg. \\
    \midrule
    PVNet\cite{peng2019pvnet}    & 25.3 & 12.7 & 4.2 & 9.4 & 19.2 & 7.7 & 13.1 \\
    Onepose\cite{sun2022onepose}  & 90.0 & 98.1 & 74.0 & 87.3 & 81.9 & 78.9 & 85.0 \\
    Ours     & 100 & 100 & 99.5 & 99.7 & 71.2 & 60.4 & \textbf{88.5} \\
    \bottomrule
  \end{tabularx}
\end{table}

\subsection{Ablation Study}

\subsubsection{Effectiveness of Scene Prior}
To validate design choices of the proposed training scheme, we decouple it into three components: multi-scenes pre-training(\textit{pre-train}), scene coordinate-based feature(\textit{coord}) and per-scene optimization(\textit{per-scene}). As shown in Table~\ref{tab:ab-finetune}, for 7Scenes our method trained on multiple scenes jointly already outperforms the strong scene coordinate regression-based baseline method DSAC*\cite{brachmann2021visual}. Per-scene optimization injects valuable scene prior to aid localization, which further boosts the accuracy significantly(last two rows). Specifically, if the scene coordinate-based feature is explicitly used as we proposed, the network learns better. From the third row, we can see that per-scene training from scratch does not achieve comparable performance as other settings. In summary, (1) multi-scenes training is important for learning general knowledge to avoid overfitting, (2) scene prior can be used to improve scene-agnostic localization, especially by adding learned scene coordinate features to the original 3D features.

% 1. 有无finetune 的对比
% 2. from scratch 和 finetune的对比
% 3. 是否基于scene coord 去finetune的对比
\begin{table}[H]  
  \caption{\textbf{Ablation study about training scheme and aggregation.}}
  \label{tab:ab-finetune}
  \centering
  \begin{tabular}{c|c|c|c||c|c}
    \toprule
    pre-train & coord & per-scene & nerf-agg & 7Scenes & Cambridge \\
    \midrule
    \cmark    & \xmark & \xmark & \xmark  & 84.8    & 79.0  \\
    \cmark    & \xmark & \xmark & \cmark  & 85.8    & 79.5  \\
    \xmark    & \cmark & \cmark & \cmark  & 81.2    & 78.3  \\
    \cmark    & \xmark & \cmark & \cmark  & 88.2    & 80.3  \\
    \cmark    & \cmark & \cmark & \cmark  & \textbf{89.5}    & \textbf{81.6} \\
    \bottomrule
  \end{tabular}
\end{table}

\subsubsection{Effectiveness of Conditional NeRF}
By design, the rendering head in our pipeline is optional. We found that rendering loss does not contribute to the matching accuracy in our experiments. However, according to Table~\ref{tab:ab-finetune}, the feature aggregation of conditional NeRF improves 3D-2D matching, especially on indoor data with more occlusions. When \textit{nerf-agg} is removed, multiview feature aggregation is by simply averaging.

\subsubsection{Effectiveness of Appearance Adaptation}
To validate the effectiveness of the appearance adaptation layer, we change the style of test images for evaluation. Specifically, experiments are conducted on Cambridge dataset, where daytime images are transferred to nighttime with the method in \cite{sakaridis2020map}. Besides the median translation error and accuracy, matching IoU(intersection-over-union between predicted and ground truth 3D-2D pairs) is also reported as the direct measurement of matching quality, considering the performance gap may be reduced by the robust estimator in PnP solver. Peak Signal-to-Noise Ratio(PSNR) is the metric to evaluate novel view synthesis. From Table~\ref{table-ab-appearance}, three conclusions can be drawn, 1) appearance changes of test images lead to a significant drop of localization accuracy(79.5 to 73.6) as well as the view synthesis quality, 2) adding stronger color jitter augmentation during training only helps improving localization in the day-to-night transfer setting but not novel view synthesis 3) localization and view synthesis performance are both better by adding the proposed appearance adaptation layer as shown in the third row of Table~\ref{table-ab-appearance}. 

\begin{table}[H]  
  \caption{\textbf{Ablation study about appearance adaptation.}}
  \label{table-ab-appearance}
  \centering
  \begin{tabular}{c|c||c|c|c|c}
    \toprule
    Color Jitter & Appearance & Error$\downarrow$ & Accuracy & IoU & PSNR \\
    \midrule
    \xmark & \xmark & 20.0 & 73.6 & 0.254 & 14.4 \\
    \cmark & \xmark & 19.1 & 75.1 & 0.286 & 15.5 \\
    \cmark & \cmark & \textbf{17.5} & \textbf{75.8} & \textbf{0.299} & \textbf{19.7} \\
    \bottomrule
  \end{tabular}
\end{table}

\section{Conclusion}
In this paper, we present a novel visual localization pipeline based on direct 3D-2D matching between the NeRF model and images. By conditioning the NeRF model on support images, the proposed method is scene agnostic. To inject scene prior, a scene coordinate-based 3D feature module for per-scene optimization is proposed, which greatly improves the per-scene performance. Moreover, we propose an appearance adaptation layer to better handle the domain gap between the query image and the conditional 3D model, which enhances the system's robustness in terms of novel view synthesis and localization. 

\bibliographystyle{plain}
\bibliography{refs}

%%%%%%%%%%%%%%%%%%%%%%%%%%%%%%%%%%%%%%%%%%%%%%%%%%%%%%%%%%%%%%%%%%%%%%%%%%%%%%%%

\end{document}